\documentclass{article}
\usepackage{arxiv}
\usepackage{float}
\usepackage[utf8]{inputenc} 
\usepackage{helvet} 
\usepackage[T1]{fontenc}   
\usepackage{hyperref}       
\usepackage{url}        
\usepackage{booktabs}
\usepackage[table,xcdraw]{xcolor}
\usepackage{booktabs}       
\usepackage{amsfonts}       
\usepackage{nicefrac}       
\usepackage{microtype}    
\usepackage{lipsum}
\usepackage{graphicx}
\usepackage{courier}
\usepackage{amsmath}
\usepackage{amssymb}
\usepackage{amsthm}
\usepackage{subfig}
\usepackage[singlelinecheck=false]{caption}
\graphicspath{ {./images/} }
\usepackage[
backend=biber,
style=apa,
sorting=nyt
]{biblatex}
\title{Towards a Fully Interpretable and More Scalable 
RSA Model for Metaphor Understanding
}
\author{
  Gaia Carenini \\
  Department of Computer Science\\
  École Normale Supérieure-PSL\\
  Paris, France \\
  \texttt{gaia.carenini@ens.psl.eu} \\
  \And
  Luca Bischetti\\
  Laboratory of Neurolinguistics and Experimental Pragmatics (NEP)\\
  Department of Humanities and Life Sciences\\ University School for Advanced Studies IUSS\\ Pavia, Italy\\
\texttt{luca.bischetti@iusspavia.it}
  \And
 Walter Schaeken\\
 Department of Psychology\\
 KU Leuven\\
 Leuven, Belgium\\
\texttt{walter.schaeken@kuleuven.be}
  \And
  Valentina Bambini\\
  Laboratory of Neurolinguistics and Experimental Pragmatics (NEP)\\
  Department of Humanities and Life Sciences\\ University School for Advanced Studies IUSS\\ Pavia, Italy\\
  \texttt{valentina.bambini@iusspavia.it} \\
}
\begin{document}
\maketitle
\begin{abstract}
The Rational Speech Act (RSA) model provides a flexible framework to model pragmatic reasoning in computational terms. However, state-of-the-art RSA models are still fairly distant from modern machine learning techniques and present a number of limitations related to their interpretability and scalability. Here, we introduce a new RSA framework for metaphor understanding that addresses these limitations by providing an explicit formula - based on the mutually shared information between the speaker and the listener - for the estimation of the communicative goal and by learning the rationality parameter  using gradient-based methods. The model was tested against 24 metaphors, not limited to the conventional \emph{John-is-a-shark} type. Results suggest an overall strong positive correlation between the distributions generated by the model and the interpretations obtained from the human behavioral data, which increased when the intended meaning capitalized on properties that were inherent to the vehicle concept. Overall, findings suggest that metaphor processing is  well captured by a typicality-based Bayesian model, even when more scalable and interpretable, opening up possible applications to other pragmatic phenomena and novel uses for increasing Large Language Models interpretability. Yet, results highlight that the more creative nuances of metaphorical meaning, not strictly encoded in the lexical concepts, are a challenging aspect for machines.
\end{abstract}
\section{Introduction}
Modeling language is a fundamental step for understanding human communication and improving human-computer interaction. To date, some domains of linguistic competence, such as syntax, have been largely investigated with formal and computational tools, whereas other areas – especially the more social ones such as pragmatics – have been usually explained in more descriptive terms. There are only a handful of studies that focus on formal and computational aspects of pragmatics (\cite{Bunt17}). Among these, the Rational Speech Act framework (RSA; \cite{FrankGoodman12}; \cite{GoodmanFrank16}) is perhaps the most noteworthy research program for investigating pragmatic reasoning and cooperative communication between interlocutors, based on the general principles of verbal interactions (\cite{Grice1991}; \cite{SperberWilson1995}). The RSA framework assumes that language production and interpretation are inherently probabilistic processes that give rise to variable choices. More precisely, the RSA approach is both a game-theoretic and decision-theoretic model that treats language use as an instance of a signaling game (\cite{Lewis1969}). Since its introduction, RSA has contributed to improving the understanding of a wide range of pragmatic phenomena, e.g., reference (\cite{FrankGoodman12,Qing2015, Heller2016, KreissDegen2020,Degen2020, Hawkins2021}), scalar implicatures (\cite{Rohde2012,Stuhlmuller2013, Bergen2016}), plural prediction (\cite{scontras2017resolving}), hyperbole (\cite{Kao2014}), and irony (\cite{Kao2015}; for a survey, see \cite{degen2023rational}).

Although the RSA framework is inherently computational and mathematically grounded, its approach is still based on classical Bayesian statistics rather than on modern machine learning or optimization techniques. For this reason, RSA suffers from high computational costs (i.e., high number of operations required to compute language predictions), poor scalability (i.e., low adaptability of the method to large data samples; \citeauthor{degen2023rational}, 2023), and out-of-domain generalization. Moreover, the use of data interpolation to obtain several model parameters amplifies such issues. In this work, we try to bridge the distance between the Rational Speech Act framework and state-of-the-art AI systems by proposing novel solutions to make the RSA model more scalable yet interpretable. As a test-ground, we focus on modeling pragmatic reasoning involved in metaphor comprehension.

Metaphor is a figurative use of language representing a radical gap between what is literally said and what is intended. Among the theoretical accounts, Relevance Theory offers a pragmatic framework for describing how metaphors are comprehended (\cite{SperberWilson1995, Carston2010}). In Relevance Theory, metaphor processing is described as a set of cognitive operations focused on inferring the speaker’s intended meaning via adjusting the denotation of the lexically encoded concepts (i.e, promoting and dropping certain concept features depending on the context), yielding an ad-hoc concept that becomes relevant for the derivation of the implicated meaning (\cite{Wilson2007}). While psycholinguistic research has provided detailed descriptions of metaphor understanding (\cite{gibbs1994poetics,Glucksberg2003,bambini2012metaphor,Holyoak2018MetaphorCA}), computational models are still struggling to fully account for the interpretative nuances of metaphors (\cite{kintsch2000metaphor,shutova2010automatic, utsumi2011computational,veale2016metaphor, Su2017}), and Artificial Intelligence and Large Language Models (LLMs) are, to date, limitedly capable of grasping metaphorical meaning (\cite{Barattieri2023,Chakrabarty2022}).  

An alternative framework for metaphor modeling, taking into account pragmatic insights too, is represented by RSA. The first RSA model for metaphor understanding was proposed by \citeauthor{Kao2014} (2014). In their formalization, the pragmatic listener assumes that the speaker chooses a metaphorical expression to maximize informativeness and consistently with her communicative goal. Their model stands out as being able to explain the use of utterances that are known to have false literal interpretations and aligns with the theoretical idea that a listener relies on communicative principles to infer what the speaker intends to communicate. In more concrete terms, Kao and colleagues proposed that when using a metaphor predicating properties of animals, the speaker aims to communicate features that are concurrently typical of an animal and relevant to humans. For example, while interpreting \emph{John is a shark}, the listener is unlikely to think that John actually has fins or swims fast, and rather assumes that the speaker aims to communicate a feature of sharks that is relevant to the referent John, such as scariness or aggressive behavior. This model was able to capture the informativeness of this type of metaphor on the basis of communicative goals, showing a strong alignment between model predictions and human data ($r=.70$). Building on this formalization, \citeauthor{Mayn2021} (2022) revisited the computational model presented by Kao and colleagues (2014) to account for the fact that, when hearing a metaphor, the listener does not simply characterize features in mere terms of presence or absence, but also infers also how typical features are in bearing metaphorical meanings. For instance, when hearing \emph{John is a shark}, the listener might interpret the speaker’s intentions to convey one or more typical features of sharks in relation to John, who possibly is a scary or aggressive human being. Mayn and Demberg proposed a graded approach representing the features of the metaphorical vehicles in terms of their degree of typicality, defined as the dimension that quantifies how much a feature belongs to the denotation of a person, an animal, or an object category. The resulting model was shown to be still able to emulate human responses with moderate accuracy ($r=.59$), while being more expressive, i.e., more informative, than in the Kao et al.'s one.

As a whole, the available RSA models can capture some relevant facets of metaphor understanding with different degrees of success, yet they present the general theoretical weaknesses of the RSA approach, as well as poor interpretability. Firstly, as concerns the latter aspect, they do not account explicitly for the role of context, whereas metaphor interpretations are inferentially derived using contextual cues (\cite{Sperber2008};  \cite{Allott12}). More precisely, no closed-form is introduced for the estimation of the probability that the pragmatic speaker has a given communicative goal, a prior that can change based on the conversational context. This compromises the overall interpretability of the model, since no instruction is given on how to compute its parameters. Secondly, they compute relevant model parameters through data interpolation encompassing huge computational costs and major shortcomings for scalability. Thirdly, the evaluation of these RSA models was limited to conventional metaphors in the \emph{X} is \emph{Y} form, with \emph{X}s (i.e., the topic of the metaphor) being male proper names and \emph{Y}s (i.e., the vehicle used metaphorically to predicate about the topic) being animals (e.g., \emph{John is a shark}).

Here, we address these weaknesses by presenting a novel RSA model that: (a) includes a closed-form for communicative goal probability distribution when the metaphor is considered in isolation, that is, where the conversational context is limited – albeit present – as induced by the topic; (b) efficiently learns – through gradient-based methods used in state-of-the-art machine learning – the rationality parameter involved in the language production actor, which we hypothesized to be invariant to the context and thus inherent to the speech actors; and (c) is tested against a set of metaphors not limited to the conventional type with proper names and animal vehicles, but including a more diverse set of topics and vehicles (e.g., as in the case of \emph{Workers are ants}).

\section{Computational Model}
We present a model that globally preserves the formalization of Kao et al. (2014) and the idea of graded typicality of Mayn \& Demberg (2022), but introduces two major modifications dealing with (i) the estimation of the parameter $\lambda$ included in the definition of the pragmatic speaker and (ii)  the mathematical expression accounting for the choice of the listener’s communicative goal in the context given by the topic of the metaphor. In what follows, we will explicitly point out similarities and differences with previous models.

Differently from Kao and colleagues (2014), we specified the topic of the \emph{X is Y} metaphors (instead of using generic proper names). We restrict the possible features of \emph{X} and \emph{Y} under consideration to a vector of size $n$: $f=[f_1,\dots, f_n]$, where, for every $i\in\{1,\dots, n\}$, $f_i$ is a number in the interval $[0,1]$, and the following condition holds $\sum_{i\in[n]}f_i=1$. As in Mayn \& Demberg (2022), this range for the entries of $f$ leads to a more fine-grained quantification of typicality with respect to the work of Kao et al., where the entries were Boolean.  As in Kao and colleagues (2014), the literal listener  will interpret the utterance \emph{X is Y} as meaning that \emph{X} is literally a member of the category \emph{Y} and has corresponding features. Formally, if $u$ is the uttered category:
\begin{equation}
L_0(c, f|u)=
\begin{cases}
    P(f|c),& \text{if } c=u\\
    0      & \text{otherwise}
\end{cases}  
\end{equation}
where $P(f|c)$ is the prior probability that a member of category $c$ has feature vector $f$.\vspace{0.2cm}\\
We assume, following Kao and colleagues (2014), that the speaker aims to highlight a specific feature of the topic. Consequently, the speaker's goal can be seen as a mapping from the complete feature space to the subset that holds relevance for them. Formally, the goal to communicate about feature $i\in\{1,\dots, n\}$ is the function $g_i(f)=f_i$.\\  
The pragmatic speaker is characterized by a utility function that assigns a utility to the pair utterance-intended meaning, based on how the literal listener interprets the utterance. Adhering to previous RSA models for metaphor, the speaker's utility is defined as the negative surprisal of the true state, given an utterance, under the listener's distribution. In this case, our attention is solely on the surprisal along the goal dimension. To achieve this, we project along the goal dimension, leading to the following function: 
\begin{equation}\label{eq: utility}
U(u|g,f)=\log\sum_{c,f'}\delta_{g(f)=g(f')}L_0(c, f'|u)
\end{equation}
The expression describing our pragmatic speaker is exactly the same as in Kao and colleagues (2014), our major improvement being that, unlike previous formalizations, our model learns the parameter $\lambda$ from data via a modern machine learning technique, namely unconstrained optimization through the conjugate gradient algorithm. Specifically, based on the utility function (Eq. \ref{eq: utility}), the pragmatic speaker $S_1$ employs a softmax decision rule, which approximates the behavior of a rational planner, to select an utterance:
\begin{equation}
    S_1(u|g,f)\sim e^{\lambda U(u|g,f)}
\end{equation}
where $\lambda$ is a parameter accounting for the rationality of the speaker. When $\lambda$ approaches infinity, $S_1$ approaches a fully rational speaker who will choose a message that maximizes $U(u|g,f)$ with probability 1. When $\lambda$ is finite, the speaker is only approximately rational, and messages that yield greater have a greater probability (although not certainty) of being used than less useful messages.

Then, we can define the pragmatic listener $L_1$ who uses Bayesian inference to derive the intended meaning based on prior knowledge and their understanding of the speaker. To determine the speaker’s intended meaning, $L_1$ considers all possible speaker’s goals and integrates them. Unlike previous models, we explicitly encode the fact that the communicative goals are considered non-uniformly by conditioning the associated probability distribution to the conversational context. To reasonably limit the scope of our work, we focused on metaphors where the conversational context is entailed by the topic of the metaphor. Hence, we define the pragmatic listener $L_1$ as:
\begin{equation}
  L_1(c,f|u)\sim P(c)P(f|c)\sum_{g} \mathcal{R}(g|t)S_1(u|g,f)
\end{equation}
where $c,f,u$, and $t$ denote respectively a category, a feature vector, an utterance, and the topic of the metaphor under discussion, and $\mathcal{R}$ is a function expressing the relevance of the goal $g$ to the topic $t$. Other parameters involved in the modeling, estimated on human data, are: a) the prior probability that the entity discussed belongs to category $c$, $P(c)$; b) the prior probability that a member of category $c$ has feature vector $f$, $P(f|c)$; and c) the parameter accounting for minimal context, $\mathcal{R}(g|t)$. This results in a direct dependence of the communicative goal estimate on the topic and such a relationship is expressed through the typicality of features obtained from a source such as human ratings.\vspace{0.2cm}\\
The key differences from previous RSA models for metaphor are the fact that $\lambda$ is assumed to be an inherent property of the speaker that can be efficiently learned through advanced optimization methods from reduced data sample, resulting in a better scalability of the model, and in the introduction of the closed formula $\mathcal{R}(g|t)$, which allows us to easily interpret how the model estimates the communicative goal starting from the conversational context given by the topic and the common background. Figure 1 illustrates the architecture of the proposed RSA model.
\begin{figure}[H]
    \centering
    \includegraphics[width=0.75\linewidth]{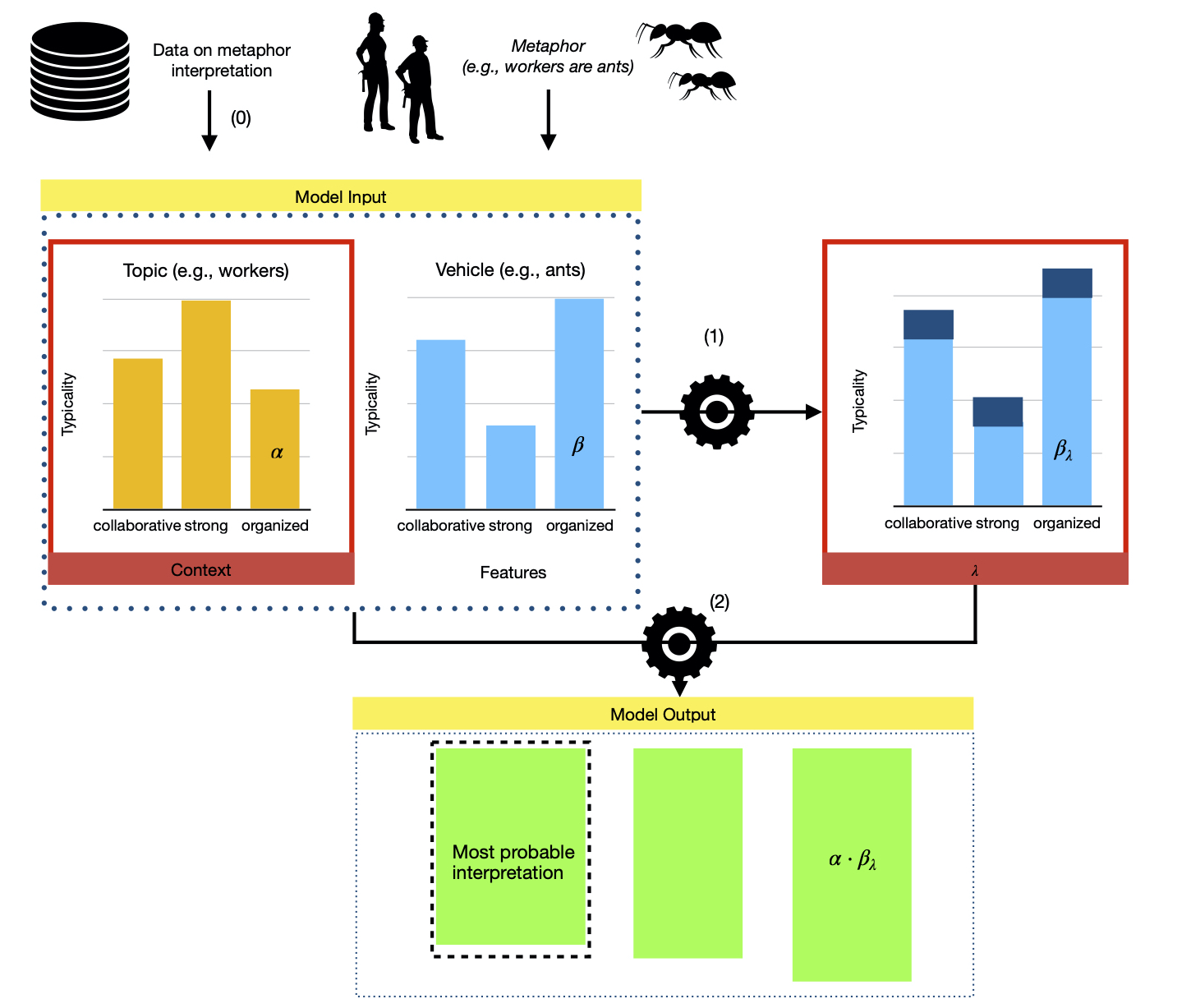}
    \caption{
    \begin{small}
Schematic representation of the RSA model for metaphor understanding. The model input is represented in the box on the top of the figure, and consists of topic-vehicle pairs (in our example, \emph{workers-ants}) and associated values of typicality for a set of features (in our case, \emph{collaborative, strong, organized}). Given this input, the model uses the parameter $\lambda$, which was previously learned from data (step 0 in the figure), in order to “stretch” the typicality values for the vehicle (e.g., to obtain $\beta_{\lambda}$) from $\beta$ (step 1). These values, which provide a proper estimator for the probability distribution over all the possible communicative goals, are combined with the typicality ones for the topic (e.g., $\alpha$; step 2) to build the final distribution over all the possible interpretations represented in the output box at the bottom of the figure.
    \end{small}
    }
    \label{fig: model}
\end{figure}
\section{Behavioural Experiments}
We conducted three experiments with human participants, mostly following Kao and colleagues (2014). Experiment 1 was carried out to elicit topic and vehicle features for a set of metaphors, accounting for a minimal context; Experiment 2 computed the conditional probabilities for topics and vehicles, while Experiment 3 collected participants’ interpretations of the metaphors. The study was approved by the Ethics Committee of the [\emph{deleted for anonymization purposes}]. Informed consent was obtained from all participants.
\subsection{Data availability statement} The study material, data, and code are stored in a repository hosted by the OSF platform \url{https://osf.io/kmv6u/?view_only=e974bb59503a430885fad5a9e1999f80}. 
\subsection{Experiment 1. Feature Elicitation}
\textbf{Materials:} We used a set of 24 \emph{Xs are Ys} metaphors, with \emph{Xs} expressing a human category and \emph{Ys} being animals or objects, selected from the Italian dataset used in \citeauthor{Canal2022} (2022). For each selected metaphor, we checked whether the implied meaning aligned with the salient semantic features of the vehicle, by using the norms in \citeauthor{McRae2005} (2005). When the implied meaning was included in the norms, the item was classified as a metaphor with vehicle-inherent properties ($N$ = 12; e.g, \emph{Dancers are swans}, with \emph{elegance} included in the norms for \emph{swan}; see Appendix Table A.1), otherwise the item was classified as a metaphor with non vehicle-inherent properties ($N$ = 12; e.g., \emph{Workers are ants}, with \emph{diligence} not included in the norms for \emph{ant}; see Appendix, Table A.2). The two sets did not differ for familiarity (metaphors with vehicle-inherent properties, $M$ = 4.40, $SD$ = 1.39; metaphors with non vehicle-inherent properties, $M$ = 4.19, $SD$ = 1.36; $\Delta M$ = 0.21, \textit{t}(22) = 0.38, \textit{p} = .710). We extracted all the topics and the vehicles of the 24 items, for a total of 48 nouns expressing human, animal, and object categories.\\
\textbf{Methods:} Fifty-eight native speakers of Italian (Age, \emph{M} = 24.98; M = 36\%, F = 59\%, other = 5\%) were administered a web-based survey via LimeSurvey. The set of 48 nouns was split into two lists, with an equal number of metaphors’ topics and vehicles, presented in a random order. Each participant was presented with one list and was asked, for each noun, to generate 3-to-5 relevant features (preferably nouns) that came to mind (e.g., \emph{workers: strength, hardness, effort}).\\ 
\textbf{Results:} We constructed a list of features for each category noun and sorted them according to their frequencies over the pool of participants. To avoid redundancy in the feature set, we used the Treccani dictionary of Italian to identify and aggregate synonyms. Next, to limit the vocabulary to a reasonable size, we extracted the features present in at least 10\% of the answers. From 4,615 observations, 59 features for the set of 48 topics and vehicles were retained (see online repository).
\subsection{Experiment 2. Typicality Elicitation}
\textbf{Materials:} We used the 48 category nouns derived from the metaphors and the 59 features extracted in Experiment 1.\\ 
\textbf{Methods:} Thirty-seven native speakers of Italian (Age, \emph{M} = 25.43; M = 24\%, F = 76\%) were administered a web-based survey on Limesurvey. The set of nouns was split into 4 lists with 12 nouns each. Participants were presented with one list and were asked, for each noun (e.g., \emph{workers}) to evaluate, on a 1-7 Likert scale, how typical each of the 59 features is (e.g., \emph{diligence, elegance, noisiness}, etc.).\\
\textbf{Results:} We derived typicality ratings for each feature for each category noun (e.g., how typical \emph{diligence} is for \emph{workers}). These values were normalized and used to estimate $P(f|c)$ and $\mathcal{R}(g|t)$.
\subsection{Experiment 3. Metaphor Understanding}
\textbf{Materials:} We used the 24 metaphors from Canal and colleagues (2022) and the 59 features extracted in Experiment 1.\\
\textbf{Methods:} Forty native speakers of Italian (Age, \emph{M} = 27.08; M = 23\%, F = 77\%) were administered a web-based survey on LimeSurvey. Each participant was presented with the full set of metaphors in a random order. For each item, participants completed a forced-choice metaphor interpretation task, selecting one among the 59 features.\\
\textbf{Results:} For each metaphor, we obtained the probability distribution over all possible interpretations by computing the cumulative popularity of each feature, subsequently normalized over the whole set of features.
\section{Model Evaluation}
The first step was computing the value of the rationality parameter, useful for obtaining adequate metaphoric interpretations from the model. For this purpose, we partitioned the 24 metaphors in a training (18 metaphors) and a test (6 metaphors) dataset, and we initialized and performed an unconstrained convex maximization of the correlation between model predictions and human ratings when restricted to the training dataset. The final value, obtained by running the conjugate gradient algorithm, was 44.43. 

The second step was assessing the performance of our model in terms of level of agreement, correlation, and dissimilarity between the metaphor interpretations returned by the algorithm and human data. Finally, to quantify the improvement provided by our model vs. other approaches, we measured the effect of introducing $\mathcal{R}(g|t)$ and the effects of learning on the overall model performance through properly designed ablation studies.

\subsection{Human vs. Model: Agreement}

We define the notion of $k$-agreement between two probability distributions to be the number of common features between the $k$ most probable ones, with the standard notion of accuracy for a predictive model corresponding to 1-agreement. In the whole set of metaphors, we obtained moderate agreement: in 11 out of 24 metaphors (46\%), the most probable feature (i.e., metaphor interpretation) from the model matched human data; in detail, 1-agreement was moderate (i.e., 58\% accuracy) for metaphors with vehicle-inherent properties and fair (i.e., 33\% accuracy) for metaphors with non vehicle-inherent properties (see Appendix, Table A.1 and A.2; Figure 2-a). Acknowledging that a metaphor may convey a range of weakly implicated meanings (\cite{Sperber2008}) and admit more than one interpretation (and thus select more than one feature), we also considered the agreement for increasing values of $k$. When $k$ = 3, we obtained a $k$-agreement of 1.37 (the model assigned an interpretation which is included in the 3 most probable ones for humans for 100\% of metaphors with vehicle-inherent properties and for 67\% of the metaphors with non vehicle-inherent properties; Appendix, Table A.1 and A.2; Figure 2-b). To better investigate the interdependencies among the features, we studied the correlation between the interpretations given by the RSA model and the human responses (Figure 3). We can notice that several clusters, i.e., groups of highly correlated features, emerged consistently across model and human data. 
\begin{figure}[H]
    \centering
    \includegraphics[width=0.90\textwidth]{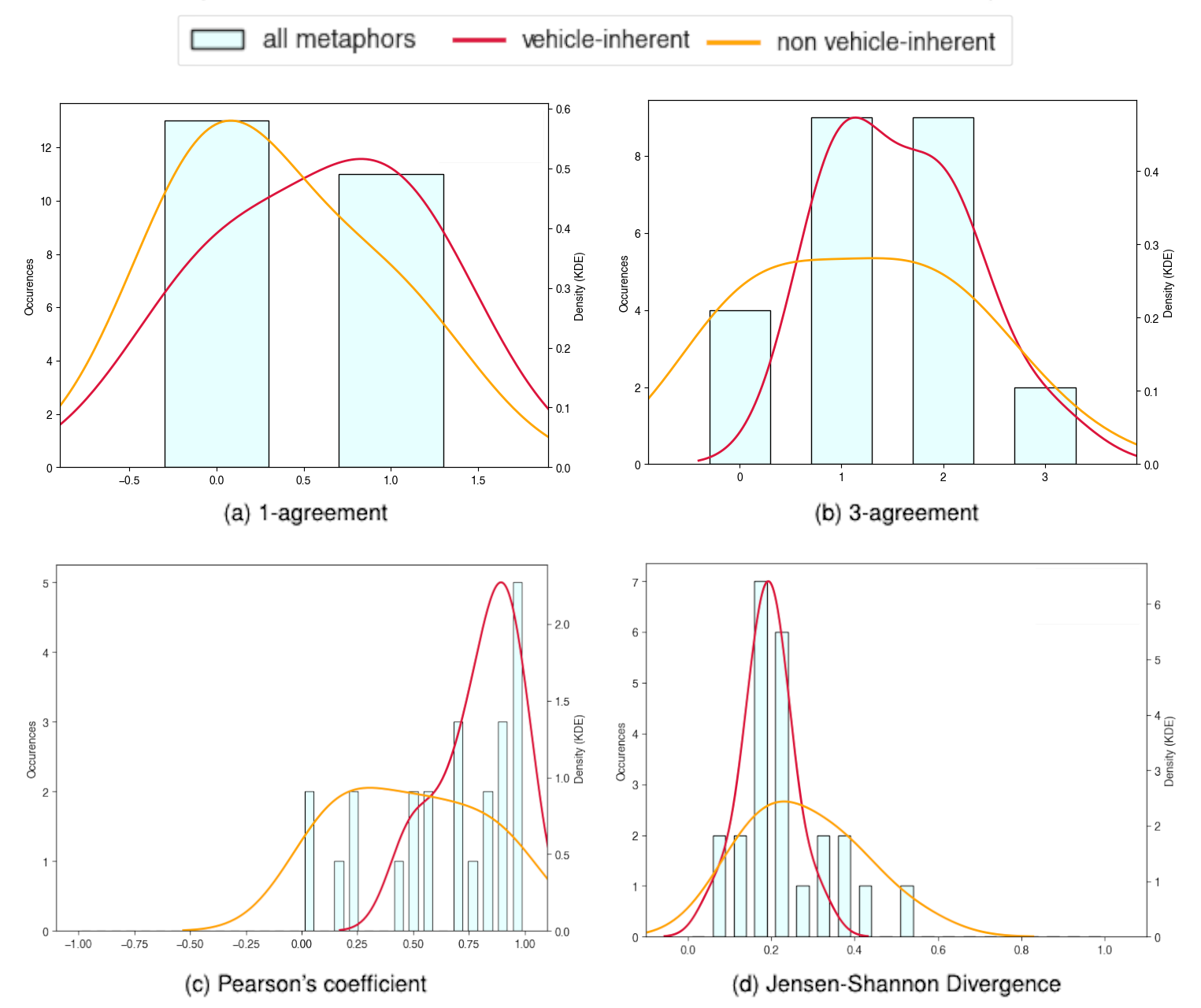}
    \caption{
    \begin{small}
       Model vs. human interpretations: $k$-agreement, Pearson’s coefficient, Jensen-Shannon Divergence. In panel (a) and (b), the histograms report the frequency of overlaps of size $0,\dots, k$ for $k=1,3$ over the $k$ most common interpretations for the model and humans. The density curves represent the $k$-agreement distinguishing metaphors with vehicle-inherent and non vehicle-inherent properties. In panel (c) and (d), the histograms report the distributions of Pearson’s correlation coefficients (c) and Jensen-Shannon Divergence. The density curves represent the coefficients distinguishing metaphors with vehicle-inherent and non vehicle-inherent properties. 
    \end{small}}
\end{figure}

\subsection{Human vs Model: Correlation and (Dis)similarity}
To further evaluate the quality of our model, we correlated the model's predictions with human interpretations of the metaphorical utterances. Given a metaphorical utterance, we computed the model’s marginal posterior probabilities for $f_i$ for $i=1,\dots,n$. We then correlated these posterior probabilities with participants’ probability ratings obtained in Experiment 3. We evaluated our model with two global indicators of (dis)similarity between human data and model-based distributions, namely Pearson’s correlation coefficient ($r$) and Jensen-Shannon Divergence ($JSD$), with the latter serving as a measure of the information shared between the model’s distribution and the observed interpretations.
The correlation quantified by the Pearson correlation coefficient was strong for the whole set ($r = .64$, $SD$ = .29), ranging from very strong for metaphors with vehicle-inherent properties ($r = .80$, $SD$ = .16) and moderate for metaphors with non vehicle-inherent properties ($r = .48$, $SD$ = .31) (Figure 2-c). The Jensen-Shannon Divergence was restrained, i.e., it described the human and the model interpretation as very similar, with $JSD = .23$ ($SD$ = .11) for the whole set, $JSD=$ .19 ($SD$ = .05) for metaphors with vehicle-inherent properties, and $JSD=$ .27 ($SD$ = .12) for metaphors with non vehicle-inherent properties (Figure 2-d). 

\begin{figure}[H]
    \centering
    \includegraphics[width=.85\linewidth]{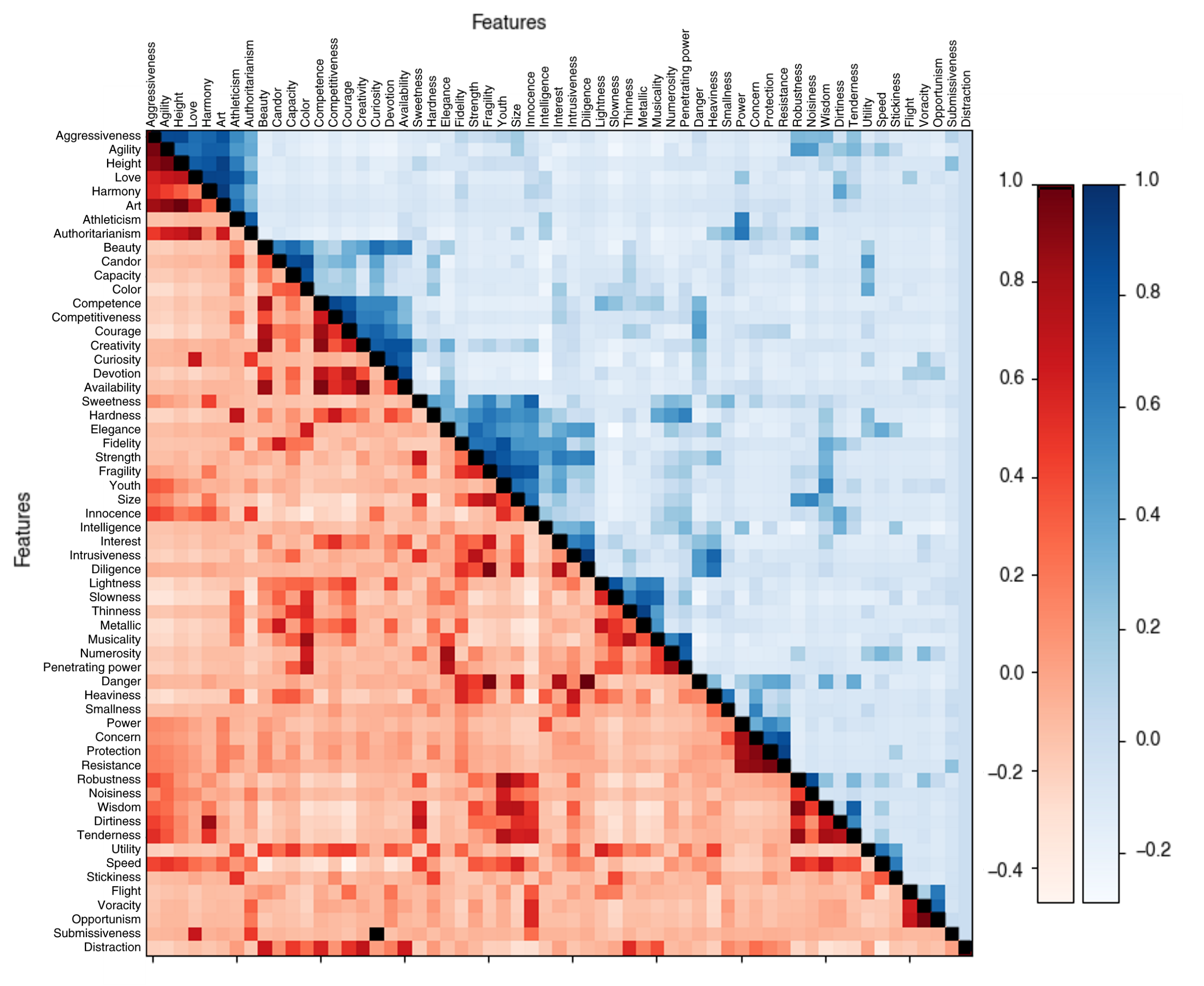}
    \caption{
    \begin{small}
      Correlations among features in metaphor interpretations in humans vs. RSA model. The table shows the correlations between pairs of features (translated from Italian into English) in the metaphor interpretations given by humans (below the diagonal) and by our model (above the diagonal). The colors of the cells represent the correlation values according to the scales on the right, (i.e., darker colors are associated with stronger correlation).   
    \end{small}
    }
    \label{fig:enter-label}
\end{figure}
\subsection{Ablation study: Effect of introducing \textbf{$\mathcal{R}(g|t)$} and learning }
Finally, we conducted an ablation study to measure the improvement induced by the modifications introduced in our model over previous formalizations. Ablation studies are a class of popular experiments within the machine learning literature consisting of systematically removing a specific component of a model to analyze its individual impact on the system's behavior. Here, we removed $\mathcal{R}(g|t)$ and performed instead a straightforward interpolation, obtaining a model in between the one we presented and the one by Kao et al. (2014). The results of this ablation study showed that removing $\mathcal{R}(g|t)$ significantly affected the correlation between model output and human ratings ($r=.55$, -14\% with respect to the previous performance $r=$ .64). 
As for the rationality parameter $\lambda$, we compared both the Pearson's correlation coefficients and the $JSD$ values obtained when $\lambda$ is estimated through interpolation (along the lines of Kao et al., 2014) to the ones when $\lambda$ is mathematically learned as in our framework. The comparison between model and human ratings remains almost unchanged, with a minimal variation of the Pearson's correlation coefficient of .02 ($r$ = .62, -3\% compared to the previous performance $r$ = .64) and a negligible change of $JSD$ of .001 ($JSD$ = .233, +0.5\% compared to the previous performance $JSD$ = .232). 

\section{Discussion}
Our work aimed to bring RSA models for metaphor closer to modern machine learning techniques while maintaining -- indeed increasing -- their interpretability and ensuring their scalability. To this aim, we introduced an explicit formula for calculating the distribution of communicative goals when the metaphor is considered in a minimal context given by the topic. Also, we learned the parameter $\lambda$, i.e., the speaker's rationality estimate, using gradient-based methods. The model’s performance showed a fairly good fit, as indicated by the overall strong positive correlation between the distributions generated by the model and the interpretations obtained from the human behavioral data, and thus has comparable performance to previous models, even when tested on a less conventional set of metaphors.

A closer look at the results allows for highlighting the strengths and weaknesses of our model. First, the model performed better in interpreting metaphors with vehicle-inherent properties. In the case of metaphors with properties that are not essential to the vehicles, typicality values are less polarized than those for metaphors expressing properties that are inherently part of the vehicle concept. While this finding suggests that polarizing typicality values improves the performance of a model, it also shows the limitations of typicality-based models in dealing with the most creative and context-based nuances of metaphorical meanings. Second, the model was more stable when considering a broader and unordered array of interpretations ($k$-agreement), indicating that it was better at capturing a set of possible meanings generated by humans rather than a unique one. A possible reason for this behavior in terms of model machinery might be linked to the ability of the model to learn correlations among features, and then resort to the whole cluster for determining the final interpretation. This might indicate the difficulty of the model to discriminate the strongest meaning among a set of alternatives.

From a theoretical perspective, our study provides further evidence for the overall appropriateness of a pragmatic model that is mainly based on typicality and suggests that metaphor interpretation is well captured by Bayesian reasoning. Specifically, when tested exclusively on the \textit{John-is-a-shark} type of metaphor, our optimized model would likely reach 100\% accuracy (as evidenced by performance on the most conventional items), outperforming previous RSA models. However, when evaluated on a more diverse set of metaphors compared to original RSA proposals, our model reached limited performance, especially in interpreting creative expressions that do not capitalize on essential properties of concepts. Hence, we can conclude that Bayesian typicality machinery works well for some types of metaphor, but it remains to be enhanced for metaphor at large, possibly via including other aspects of metaphor comprehension, perhaps dealing with mindreading and sensory-based experience (\cite{Barattieri2023}) or its emotional impact and the complexity of the pragmatic message (\cite{GibbsCog, Holyoak2018MetaphorCA}).

At the technical level, our study has two main implications. First, it supports the possibility of introducing machine learning into the core of RSA models while maintaining its ability to emulate human pragmatic reasoning. In particular, we showed that using machine learning techniques improves the scalability of state-of-the-art RSA models, one of their most important limitations (\cite{degen2023rational}), while maintaining and improving the overall interpretability of the entire framework. To the best of our knowledge, this is the first work to suggest the possibility of deriving parameters of RSA from learning techniques, and we strongly believe that this approach can be easily generalized from metaphor to other phenomena. Moreover, machine learning optimization could be used to design novel algorithms for metaphor comprehension, able to deal with more creative and less encoded aspects of meaning (\cite{twomey2023creativity, franceschelli2023creativity, orwig2024creativity}). Second, our study might offer a key to unveil the underlying mechanisms of the last generation of large language models (LLMs), which have shown surprisingly good linguistic skills but still weaknesses in metaphor and pragmatic (\cite{pedinotti-etal-2021-howling, Barattieri2023, Borghi, hu-etal-2023-fine}) and limited understanding at large (\cite{Dentella, Moro, Len}). In particular, preliminary evidence of a strong correlation between metaphor interpretation given by our RSA model and GPT-XL (\cite{GPT-Us}), suggests that RSA mechanisms might be embedded in LLMs.

To conclude, the RSA framework seems to be a promising way to model metaphorical reasoning, at the interface between classic pragmatic theory and the most modern machine learning techniques. What remains difficult to learn are the most nuanced, creative and emotional aspects of metaphorical meaning, which are challenging for both machines, as well as for humans. 

\newpage
\printbibliography
\newpage 
\section*{Appendix}
\begin{center}
\renewcommand{\tablename}{Table A.}
\begin{table}[H]
\caption*{Table A.1: Interpretations of metaphors with vehicle-inherent properties.}  
\vspace{0.5cm}
\begin{tabular}{@{}lllll@{}}
\rowcolor[HTML]{CED7E7} 
\textbf{Metaphor}                & \textbf{Human Interpretation}                 & \textbf{RSA model Interpretation}                 & \textbf{1-agreement}
& \textbf{3-agreement} \\
\rowcolor[HTML]{E8ECF3} 
\begin{tabular}[c]{@{}l@{}}\emph{I ballerini sono cigni.} \\ Dancers are  swans.\end{tabular}            & \begin{tabular}[c]{@{}l@{}}Elegance, \\ Lightness, \\ Beauty\end{tabular}        & \begin{tabular}[c]{@{}l@{}}Elegance, \\ Harmony, \\ Lightness\end{tabular}          & 1           & 2           \\
\rowcolor[HTML]{CED7E7} 
\begin{tabular}[c]{@{}l@{}}\emph{Gli anziani sono lumache.} \\ The elderly are snails.\end{tabular}      & \begin{tabular}[c]{@{}l@{}}Slowness, \\ Tenderness, \\ Power\end{tabular}        & \begin{tabular}[c]{@{}l@{}}Slowness, \\ Fragility, \\ Stickiness\end{tabular}       & 1           & 1           \\
\rowcolor[HTML]{E8ECF3} 
\begin{tabular}[c]{@{}l@{}}\emph{I ciclisti sono razzi.}\\ Cyclist are rockets.\end{tabular}             & \begin{tabular}[c]{@{}l@{}}Speed,\\ Athleticism, \\ Opportunism\end{tabular}     & \begin{tabular}[c]{@{}l@{}}Speed,\\ Power,\\ Resistance\end{tabular}                & 1           & 1           \\
\rowcolor[HTML]{CED7E7} 
\begin{tabular}[c]{@{}l@{}}\emph{I muratori sono rocce.} \\ Masons are rocks.\end{tabular}               & \begin{tabular}[c]{@{}l@{}}Power, \\ Robustness, \\ Hardness\end{tabular}        & \begin{tabular}[c]{@{}l@{}}Hardness,\\ Robustness, \\ Heaviness\end{tabular}        & 0           & 2           \\
\rowcolor[HTML]{E8ECF3} 
\begin{tabular}[c]{@{}l@{}}\emph{I corridori sono lepri.} \\ Runners are hares.\end{tabular}             & \begin{tabular}[c]{@{}l@{}}Speed, \\ Agility, \\ Athleticism\end{tabular}        & \begin{tabular}[c]{@{}l@{}}Athleticism, \\ Agility, \\ Speed\end{tabular}           & 0           & 3           \\
\rowcolor[HTML]{CED7E7} 
\begin{tabular}[c]{@{}l@{}}\emph{I rugbisti sono tori.} \\ Rugby players are bulls.\end{tabular}         & \begin{tabular}[c]{@{}l@{}}Power,\\ Robustness,\\ Strength\end{tabular}          & \begin{tabular}[c]{@{}l@{}}Strength,\\ Aggressivness, \\ Competitivity\end{tabular} & 0           & 1           \\
\rowcolor[HTML]{E8ECF3} 
\begin{tabular}[c]{@{}l@{}}\emph{I cantanti sono usignoli.} \\ Singers are nightingales.\end{tabular}    & \begin{tabular}[c]{@{}l@{}}Musicality,\\ Harmony, \\ Sweetness\end{tabular}      & \begin{tabular}[c]{@{}l@{}}Musicality, \\ Harmony, \\ Lightness\end{tabular}        & 1           & 2           \\
\rowcolor[HTML]{CED7E7} 
\begin{tabular}[c]{@{}l@{}}\emph{I papà sono ombrelli.} \\ Dads are ombrellas.\end{tabular}              & \begin{tabular}[c]{@{}l@{}}Protection, \\ Love, \\ Concern\end{tabular}          & \begin{tabular}[c]{@{}l@{}}Protection,\\ Usefulness,\\ Resistance\end{tabular}      & 1           & 1           \\
\rowcolor[HTML]{E8ECF3} 
\begin{tabular}[c]{@{}l@{}}\emph{I genitori sono scudi.} \\ Parents are shields.\end{tabular}            & \begin{tabular}[c]{@{}l@{}}Protection, \\ Resistance, \\ Robustness\end{tabular} & \begin{tabular}[c]{@{}l@{}}Protection,\\ Braveness,\\ Hardness\end{tabular}         & 1           & 1           \\
\rowcolor[HTML]{CED7E7} 
\begin{tabular}[c]{@{}l@{}}\emph{I giocatori sono elefanti.} \\ Players are elephants.\end{tabular}      & \begin{tabular}[c]{@{}l@{}}Heaviness, \\ Robustness, \\ Size\end{tabular}        & \begin{tabular}[c]{@{}l@{}}Height, \\ Strength,\\ Size\end{tabular}                 & 0           & 1           \\
\rowcolor[HTML]{E8ECF3} 
\begin{tabular}[c]{@{}l@{}}\emph{Le indossatrici sono bambole.}\\ Models are dolls.                         \end{tabular} &
\begin{tabular}[c]{@{}l@{}}Beauty,\\ Elegance,\\ Submissiveness\end{tabular}     & \begin{tabular}[c]{@{}l@{}}Youth,\\ Beauty, \\ Elegance\end{tabular}                & 0           & 2           \\
\rowcolor[HTML]{CED7E7} 
\begin{tabular}[c]{@{}l@{}}\emph{Gli scalatori sono scoiattoli.} \\ Climbers are squirrels.\end{tabular} & \begin{tabular}[c]{@{}l@{}}Agility, \\ Athleticism,\\ Harmony\end{tabular}       & \begin{tabular}[c]{@{}l@{}}Agility, \\ Athleticism,\\ Speed\end{tabular}            & 1           & 2   
\end{tabular}
\caption*{\emph{Note.} The table includes the set of metaphors with vehicle-inherent properties used in the experiment, in Italian and in the English translation (first column). For each metaphor, we include the three most probable interpretations provided by humans (second column) and by the RSA model (third column), translated into English. Moreover, we provide the value of the $k$-agreement for $k=1$ and $k=3$. }
\label{tab: vehicle-inherent-int}
\end{table}
\renewcommand{\tablename}{Table A.2}
\begin{table}[ht!]
\vspace{-6cm}
\caption*{Table A.2: Interpretations of metaphors with non vehicle-inherent properties.}
\vspace{0.5cm}
\begin{tabular}{@{}lllll@{}}
\rowcolor[HTML]{CED7E7} 
\textbf{Metaphor}                                  & \textbf{Human Interpretation}                   & \textbf{RSA model Interpretation}              
 & \textbf{1-agreement} & \textbf{3-agreement} \\
\rowcolor[HTML]{E8ECF3} 
\begin{tabular}[c]{@{}l@{}}\emph{I credenti sono greggi.} \\ Believers are flocks.\end{tabular}            & \begin{tabular}[c]{@{}l@{}}Submissiveness, \\ Devotion,\\ Numerosity\end{tabular}        & \begin{tabular}[c]{@{}l@{}}Numerosity, \\ Fidelity,\\ Devotion\end{tabular}            & 0           & 2           \\
\rowcolor[HTML]{CED7E7} 
\begin{tabular}[c]{@{}l@{}}\emph{I buttafuori sono armadi.} \\ Bouncers are closets.\end{tabular}          & \begin{tabular}[c]{@{}l@{}}Robustness,\\ Size, \\ Height\end{tabular}                    & \begin{tabular}[c]{@{}l@{}}Height, \\ Size, \\ Robustness\end{tabular}                 & 0           & 3           \\
\rowcolor[HTML]{E8ECF3} 
\begin{tabular}[c]{@{}l@{}}\emph{I fanciulli sono agnelli.}  \\ Children are lambs.\end{tabular}           & \begin{tabular}[c]{@{}l@{}}Innocence,\\ Tenderness, \\ Candor\end{tabular}               & \begin{tabular}[c]{@{}l@{}}Innocence,\\ Tenderness, \\ Fragility\end{tabular}          & 1           & 2           \\
\rowcolor[HTML]{CED7E7} 
\begin{tabular}[c]{@{}l@{}}\emph{I capi ufficio sono iene.} \\ Office managers are hyenas.\end{tabular}    & \begin{tabular}[c]{@{}l@{}}Aggressiveness,\\ Authority, \\ Opportunism\end{tabular}      & \begin{tabular}[c]{@{}l@{}}Aggressiveness,\\ Opportunism, \\ Intelligence\end{tabular} & 1           & 2           \\
\rowcolor[HTML]{E8ECF3} 
\begin{tabular}[c]{@{}l@{}}\emph{I giornalisti sono avvoltoi.} \\ Journalists are vultures.\end{tabular}   & \begin{tabular}[c]{@{}l@{}}Opportunism,\\ Intrusiveness,\\ Competitiveness\end{tabular}  & \begin{tabular}[c]{@{}l@{}}Aggressiveness,\\ Opportunism, \\ Voracity\end{tabular}     & 0           & 2           \\
\rowcolor[HTML]{CED7E7} 
\begin{tabular}[c]{@{}l@{}}\emph{I maestri sono libri.}   \\ Teachers are books.\end{tabular}              & \begin{tabular}[c]{@{}l@{}}Wisdom, \\ Competence, \\ Intelligence\end{tabular}           & \begin{tabular}[c]{@{}l@{}}Wisdom, \\ Interest,\\ Creativity\end{tabular}              & 1           & 1           \\
\rowcolor[HTML]{E8ECF3} 
\begin{tabular}[c]{@{}l@{}}\emph{Le mogli sono martelli.} \\ Wives are hammers.\end{tabular}               & \begin{tabular}[c]{@{}l@{}}Heaviness,\\ Intrusiveness, \\ Noiseness\end{tabular}         & \begin{tabular}[c]{@{}l@{}}Strength, \\ Power, \\ Resistance\end{tabular}              & 0           & 0           \\
\rowcolor[HTML]{CED7E7} 
\begin{tabular}[c]{@{}l@{}} \emph{I filosofi sono aeroplani.}\\ Philosophers are airplanes. \end{tabular}              & \begin{tabular}[c]{@{}l@{}}Creativity, \\ Flight, \\ Wisdom\end{tabular}                 & \begin{tabular}[c]{@{}l@{}}Size, \\ Power,\\ Competence\end{tabular}                   & 0           & 0           \\
\rowcolor[HTML]{E8ECF3} 
\begin{tabular}[c]{@{}l@{}}\emph{Le nuore sono trapani.} \\ Daughters-in-law are drills.\end{tabular}      & \begin{tabular}[c]{@{}l@{}}Intrusiveness,\\ Heaviness, \\ Penetrating power\end{tabular} & \begin{tabular}[c]{@{}l@{}}Noisiness,\\ Strength,\\ Hardness\end{tabular}              & 0           & 0           \\
\rowcolor[HTML]{CED7E7} 
\begin{tabular}[c]{@{}l@{}}\emph{Gli operai sono formiche}. \\ Workers are ants.\end{tabular}              & \begin{tabular}[c]{@{}l@{}}Diligence,\\ Numerosity, \\ Slowness\end{tabular}             & \begin{tabular}[c]{@{}l@{}}Diligence,\\ Numerosity, \\ Competence\end{tabular}         & 1           & 2           \\
\rowcolor[HTML]{E8ECF3} 
\begin{tabular}[c]{@{}l@{}}\emph{I cuochi sono mongolfiere}. \\ Cooks are hot-air balloons.\end{tabular}   & \begin{tabular}[c]{@{}l@{}}Creativity, \\ Robustness, \\ Size\end{tabular}               & \begin{tabular}[c]{@{}l@{}}Competence,\\ Height, \\ Curiosity\end{tabular}             & 0           & 0           \\
\rowcolor[HTML]{CED7E7} 
\begin{tabular}[c]{@{}l@{}}\emph{Gli impiegati sono zerbini}. \\ Office workers are doormats.\end{tabular} & \begin{tabular}[c]{@{}l@{}}Submissiveness, \\ Opportunism,\\ Smallness\end{tabular}       & \begin{tabular}[c]{@{}l@{}}Usefulness, \\ Submissiveness, \\ Availability\end{tabular} & 0           & 1          
\end{tabular}
\caption*{\emph{Note.} The table includes the set of metaphors with non vehicle-inherent properties used in the experiment, in Italian and in the English translation (first column). For each metaphor, we include the three most probable interpretations provided by humans (second column) and by the RSA model (third column), translated into English. Moreover, we provide the value of the $k$-agreement for $k=1$ and $k=3$.}
\end{table}
\label{tab: emergent-int}
\end{center}

\end{document}